\pdfoutput=1

\documentclass[11pt]{article}

\usepackage{acl}

\usepackage{times}
\usepackage{latexsym}

\usepackage{graphicx}
\usepackage{booktabs}
\usepackage{siunitx}

\usepackage{soul}

\usepackage[T1]{fontenc}

\usepackage[utf8]{inputenc}

\usepackage{microtype}
\usepackage{tabularx}
\title{Re-examining Sexism and Misogyny Classification \\ with Annotator Attitudes}

\author{Aiqi Jiang$^{1,*}$ \and  Nikolas Vitsakis$^{1,}$\thanks{\ \ These authors contributed equally.} \and Tanvi Dinkar$^{1}$  \\
        \and \textbf{Gavin Abercrombie}$^{1}$ \and \textbf{Ioannis Konstas}$^{1}$\\
        $^{1}$The Interaction Lab, Heriot-Watt University \\
        \texttt{\{a.jiang, nv2006, t.dinkar, g.abercrombie, i.konstas\}@hw.ac.uk} }

\begin{document}
\maketitle
\begin{abstract}
Gender-Based Violence (GBV) is an increasing problem online, but existing datasets fail to capture the plurality of possible annotator perspectives or ensure the representation of affected groups.
We revisit two important stages in the moderation pipeline for GBV: (1) manual data labelling; and (2) automated classification. 

For (1), we examine two datasets to investigate the relationship between annotator identities and attitudes and the responses they give to two GBV labelling tasks.
To this end, we collect demographic and attitudinal information from crowd-sourced annotators using three validated surveys from Social Psychology.
We find that higher Right Wing Authoritarianism scores are associated with a higher propensity to label text as sexist, while for Social Dominance Orientation and Neosexist Attitudes, higher scores are associated with a negative tendency to do so.

For (2), we conduct classification experiments using Large Language Models and five prompting strategies, including infusing prompts with annotator information.
We find:
(i) annotator attitudes affect the ability of classifiers to predict their labels;
(ii) including attitudinal information can boost performance when we use well-structured brief annotator descriptions;
and (iii) models struggle to reflect the increased complexity and imbalanced classes of the new label sets.\footnote{Data and code are available at \url{https://github.com/HWU-NLP/GBV-attitudes}.}

\end{abstract}

\paragraph{{\color{red}Content Warning:}} {\color{red}This document includes examples of harmful and offensive language. These are found in the Appendices.}

\section{Introduction}

Gender-Based Violence (GBV) is an increasing problem in online spaces, affecting around half of all
women and targeting those from marginalised groups in  particular~\citep{glitch-ewaw-2020-ripple,parikh-etal-2019-multi}, resulting in women often feeling uncomfortable online 
\citep{stevens2024women}.

To counter this, there have been 
attempts to facilitate content moderation
using natural language processing (NLP) methods to automatically identify 
misogynistic language. As a result, there now exist several 
datasets designed for supervised classification of various forms of GBV.
However, 
\citet{abercrombie-etal-2023-resources} identified several weaknesses in approaches to the creation of corpora for this task.
One prominent shortcoming has been the lack of representation in the labelled data of people's different points of view, particularly of 
those with 
minoritised identities 
who are 
best placed to recognise GBV.

To fill this gap, we revisit the task of classifying online text following \emph{strongly perspectivist} data practices~\cite{nlperspectives-2023-perspectivist,cabitza-etal-2023-toward},
which aim to preserve labels provided by multiple annotators in the collection and modelling of data.
We re-annotate two recent datasets, namely Explainable Detection of Sexism (\texttt{EDOS})~\citep{kirk-etal-2023-semeval},
and Detection of Online Misogyny (\texttt{DOM})~\cite{guest-etal-2021-expert}, this time with (1) multiple ratings per item; and (2) demographic and attitudinal information about the annotators, which we maintain throughout the classification pipeline. 

Prior work by \citet{davani-etal-2023-disentangling} that also collected attitudinal survey data 
from annotators
attempts to capture morality via the Moral Foundations Questionnaire (MFQ) \cite{GRAHAM201355} as a predictor of the perception of offensiveness in toxic language. 
However, due to 
criticism of the MFQ 
regarding poor internal consistency \cite{kivikangas2021moral}, we look towards other factors 
that influence individuals' responses,
with evidence from social psychology and sociology pointing towards the constructs of \emph{right wing authoritarianism} (RWA), \emph{social dominance orientation} (SDO)
and \emph{Hostile Neosexism} (HN)
\footnote{For more details on the \emph{RWA}, \emph{SDO} and \emph{HN} measures, please refer to \autoref{background_beliefs}.}
\cite{altemeyer1983right,pratto1994social,chulvi-etal-2023-social} as potentially relevant towards understanding the link between attitudes and GBV-related behaviour.

We extend a pilot study by \citet{abercrombie-etal-2024-revisiting} to explore the link between these attitudes and annotating behaviours on our re-annotated dataset, and find that annotators with higher
propensity toward RWA are more likely to label text as sexist, 
possibly due to its association with benevolent sexist attitudes \citep{de2022understanding}.\footnote{I.e.  ``Attitudes
towards women that seem subjectively positive but are actually discriminatory'' \citep{chulvi-etal-2023-social}.}
In contrast, we find that those with a higher propensity toward SDO and HN -- both associated with hostile sexism \citep{la2020social,chulvi-etal-2023-social} -- 
are less likely to label items as sexist, possibly due to the text aligning with internalised beliefs.

\begin{figure*}[ht!]
  \centering
  \includegraphics[width=0.9\linewidth]{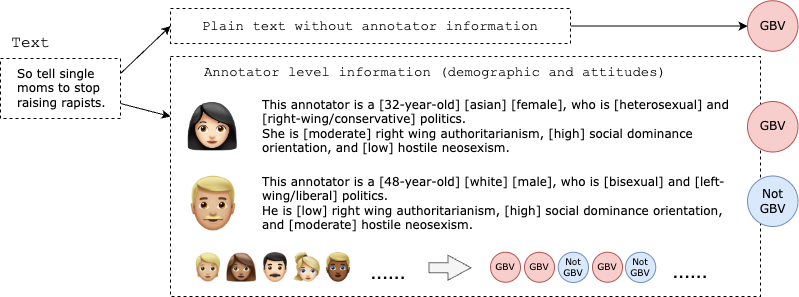}
  \caption{Two prompting paradigms under which models may assign different labels to the same text: Items are entered as plain text, or the prompt is enriched with socio-demographic and attitudinal information about annotators.} 
  \label{fig:anno}
\end{figure*}

While 
the datasets we re-annotated 
were originally conceived of for the classification of single `gold standard' aggregated labels, 
we 
aim to represent diverse perspectives in
predicting individual annotator labels 
\citep{leonardelli-etal-2023-semeval}. To better study the effect of including annotator attitudes as input to a classification task, we conduct a large set of instruction-based zero-shot, few-shot (in-context learning; ICL), and fine-tuning experiments with four open-source Large Language Models (LLMs), namely \texttt{Flan-T5} \cite{chung2022scaling}, \mbox{\texttt{Llama 2}} \cite{touvron2023llama2}, \texttt{Llama 3} \cite{meta2024llama3}, and \texttt{Mistral} \citep{jiang2023mistral}. Following \citet{fleisig-etal-2023-majority} we experiment with different prompt templates to better incorporate the annotator information (shown in \autoref{fig:anno}).
We find that ICL works 17\% and 26\% better than the majority baseline for majority vote and individual annotator tasks respectively, and fine-tuning LLMs performs 31\% better when predicting individual labels per annotator.
The best way to incorporate attitudinal data for annotators is to include well-structured brief annotator descriptions about demographics and attitudes but exclude demonstrations.
Our experimental results also indicate that models are biased towards annotators' attitudes.
\looseness=-1




\section{Background} \label{sec:background}

\paragraph{The GBV Framework} We follow \citet{abercrombie-etal-2023-resources} in adopting this framework and the term GBV as a class label. 
It encompasses phenomena such as sexism, misogyny, and violence against women and girls---although it also recognises that people of all genders are affected by GBV.

\paragraph{Annotator Variability and Perspectivist Data Practices}
While labels collected for supervised classification have traditionally been aggregated to a single `gold' or `ground truth' label for each item, recent work has recognised that this can lead to the erasure of minoritised voices. 
This occurs by either
hindering the ability of classifiers to recognise subtle and implicit forms of abuse, 
or by creating a prediction bias in the classifiers -- e.g. in the form of harmful stereotypes -- against historically minoritised voices \citep{davani-etal-2023-hate}.
\emph{Standpoint theory}~\citep{harding-1991-whose} contends that only people with relevant lived experiences are able to recognise subtle, implicit abuse such as stereotypes and micro-aggressions.
According to the \emph{matrix of domination}~\citep{collins2002black}, this experience likely results from sharing intersectional social categorisations with the intended targets of the abuse.

There is now a growing recognition of the need to collect, retain, and distribute labels provided by multiple annotators, and this has been adopted across a range of NLP tasks~\citep[for an extensive list, see][]{plank-2022-problem}.
This is particularly so for controversial tasks such as identification of abusive or toxic language, in which annotator variation may be caused by differences of opinion or ideology~\citep[e.g.][]{akhtar2021opinions,almanea-poesio-2022-armis,cercas-curry-etal-2021-convabuse,leonardelli-etal-2021-agreeing}.
\emph{Strong Perspectivism} aims to preserve this variation through modelling, classification, and evaluation~\citep{cabitza-etal-2023-toward}.\footnote{
For further background, see the Perspectivist Data Manifesto at \url{https://pdai.info/}}

\paragraph{Beliefs and Attitudes}
\label{background_beliefs}

We ground our approach to the analysis of annotator beliefs in the Dual Process Motivational Model of Ideology and Prejudice~\citep{duckitt2001dual,duckitt2009dual}.
This links sociopolitical and ideological attitudes to prejudice captured by three related constructs: 
Right Wing Authoritarianism (RWA), Social Dominance Orientation (SDO), and Hostile Neosexism (HN).
RWA explains propensity towards cultural conservatism and traditionalism-related beliefs \citep{altemeyer1983right,feather2012values, van2019religiosity}, 
while SDO explains favourable views towards social hierarchies of power, where inequality between groups is seen as inevitable or even natural \citep{christopher2008social, pratto1994social,jagayat2021cyber}.
HN is characterised by continued discrimination against women, denial of women's demands, opposition to policies aimed at improving women's social status, and the belief that feminist-driven changes unfairly disadvantage men~\citep{tougas1995neosexism,swim1995sexism,chulvi-etal-2023-social}.

These constructs have been extensively assessed and found to be strongly related. 
They
explain different forms of sexism and gender-based discrimination. 
RWA is linked to
benevolent sexism, that is attitudes that force women into traditional predefined roles (e.g., being a mother) that seem superficially advantageous but are, in reality, marginalising and disempowering~\citep{de2022understanding}. 
SDO correlates with hostile sexism, and pertains to beliefs in deterministic gender imbalances justifying male dominance through disparaging characterisations of women \citep{de2022understanding,la2020social}. 
Finally, HN is primarily associated with hostile sexist and anti-feminist attitudes in particular \citep{chulvi-etal-2023-social,off2023complexities}.


These constructs have been widely used to explain gender-based discrimination through both offline~\citep{christopher2008social,perez2021predicts,patev2019hostile,chulvi-etal-2023-social} and online~\citep{jagayat2021cyber} contexts, have been validated across cultures \citep{ccetiner2021prejudice,de2022understanding}, and previously used to explain that such beliefs transcend demographic identities \citep{renstrom2023exploring}.

\section{Related Work}


\paragraph{GBV Datasets}
\citet{abercrombie-etal-2023-resources} systematically reviewed resources for automated detection of GBV, finding a small number of datasets that contain theoretical underpinnings~\citep[e.g.][]{samory2021call,jha2017does}. 
We select two of these datasets~\citep{kirk-etal-2023-semeval,guest-etal-2021-expert} for reannotation.

\paragraph{Annotator Characteristics}
A number of NLP studies have attempted to 
use annotators'
demographic characteristics 
as predictors of their responses to items~\citep[e.g.][]{akhtar2021opinions,dutta2023modeling,gordon-etal-2022-jury,goyal-etal-2022-is, larimore-etal-2021-reconsidering,pei-jurgens-2023-annotator}. 
However, it has repeatedly been shown that demographic characteristics do not predict annotator behaviour at the individual level~\citep{beck-etal-2024-sensitivity,biester-etal-2022-analyzing,hwang-etal-2023-aligning,orlikowski-etal-2023-ecological,beck-etal-2024-sensitivity}.

Recent studies have therefore attempted to uncover annotators' \emph{social attitudes} and relate these to their responses. 
\citet{sap-etal-2022-annotators} 
found that crowd workers with racist beliefs were less likely to consider anti-Black language as toxic.
While they conducted two annotation experiments, one with many annotators but few items, and another with fewer annotators but more items, our data collection aims at both breadth and depth.
\citet{hettiachchi2023crowd} measured crowd workers' responses 
to a misogynistic language labelling task, 
and surveyed their moral attitudes (in addition to demographic and personality-type information).
They found that higher \emph{moral integrity} and lower \emph{benevolent sexism} scores correlated with label agreement with expert annotators.
\citet{davani-etal-2023-disentangling} found that while cross-cultural differences exist, individual moral values significantly influence annotators' response to perceived offensiveness levels. 
Hence, we seek to explore the relationship between demographics, social attitudes, and crowd-sourced responses to GBV identification tasks.



\paragraph{Modelling Multiple Perspectives} 
Previously, research on modelling with label variation focused on using disagreements to inform improved prediction of a single aggregated label~\citep[see][for a survey]{uma-etal-2021-learning}.
More recent work has attempted to preserve these variations at inference.
For example, \citet{cercas-curry-etal-2021-convabuse} and \citet{davani-etal-2022-dealing} predicted each annotator's responses to abusive language identification tasks, the latter using multi-task learning.
The SEMEVAL shared task on learning with disagreement (Le-Wi-Di)~\citep{leonardelli-etal-2023-semeval} explicitly attempted to focus the field on attention to levels of disagreement between annotators.
This drew several approaches including that of \citet{vitsakis-etal-2023-ilab}, who focused on preserving the full range of points of view at inference at the expense of overall classification performance.\looseness=-1

\paragraph{Toxic Language Detection with LLMs}


With the recent explosion in the use of LLMs, there has been a paradigm shift in approaches to the identification of phenomena such as toxic language as researchers have shifted from training models from scratch~\cite[e.g.][]{Davidson_Warmsley_Macy_Weber_2017,JIANG2022100182} or fine-tuning pre-trained models~\cite[e.g.][]{caselli-etal-2020-feel,cercas-curry-etal-2021-convabuse} to harnessing the power of ICL. Classification is turned into a single- or few-word generation task of the target label, merely by providing a few, or even no, specific examples as in input to the model in the form of an instruction or ``prompt'' \citep{plaza-del-arco-etal-2023-respectful,roy-etal-2023-probing,pendzel2023generative,hartvigsen-etal-2022-toxigen,sen-etal-2023-people,ziems-etal-2024-can}. This is particularly appealing given the time, effort, and cost of collecting large-scale datasets with a large pool of annotators. To that end, we benchmark the new version of the dataset and its additional labels, and examine the ability of state-of-the-art systems to recognise GBV.
\looseness=-1











\section{Data Collection}
\label{sec:datasets}

We selected the test sets and a subsection of the training sets of two 
previously published datasets: Explainable Detection of Sexism (\texttt{EDOS})~\citep{kirk-etal-2023-semeval},
and Detection of Online Misogyny (\texttt{DOM})~\cite{guest-etal-2021-expert}.
We chose these
as (1) \citet{abercrombie-etal-2023-resources} had identified them as among the resources most thoroughly grounded in social science theory; (2) they are English language datasets, the language of our stakeholder partners, with whom we are co-designing GBV-mitigation tools under the framework of participatory design; and (3) the textual data is from two different platforms, providing an opportunity for \mbox{cross-(sub-)domain} comparison.

Pre-processing consisted solely of filtering out any items which included images. 
We leave annotations of multi-media items for future work.
This left 
3,896
items, of which we re-annotated a random selection of 1,000 from the test sets for evaluation and 600 from the training sets for fine-tuning.\footnote{We maintained the 3:1 size ratio between \texttt{EDOS} and \texttt{DOM} of the original datasets.}
Table \ref{tab:classes} shows a comparison between the original and new label distributions, with the new labels determined by majority vote.
\looseness=-1

\vspace{-5pt}
\begin{table}[ht!]
    \centering
    \small
    
    \begin{tabular}{c|lc|c}
        \textbf{Dataset} & \textbf{Label} & \textbf{\#Original} & \textbf{\#New} \\
        \hline
    \texttt{EDOS}    & \emph{Sexist} & $299$ & $406$ \\
        & \emph{Not sexist} & $901$ & $794$  \\ \hline
    \texttt{DOM}    & \emph{Misogynistic} & $47$ & $97$  \\
        & \emph{Nonmisogynistic} & $353$ & $303$  \\ \hline
    Ours    & \emph{GBV} & $346$ & $503$ \\
        & \emph{Not GBV} &  $1254$ & $1097$ \\ 
    \end{tabular}
    \caption{Label distributions in the datasets. ``\#Original'' represents the distribution of labels from original data sources, and ``\#New'' represents the distribution of our re-annotated labels, determined by majority vote.}

    \label{tab:classes}
\end{table}
\vspace{-10pt}



As the Amazon Mechanical Turk (MTurk) crowd-sourcing platform is 
widely
used to collect annotations and personal information for sensitive tasks \citep{sap2019social,kumar2021designing},
we recruited 43 annotators on MTurk (19 women and 24 men with a mean age of 38, see \autoref{app:data_statement} for a full Data Statement with detailed annotator information).
To ensure attentive participation, we recruited only workers with 
$\geq500$ completed tasks and a $\geq98\%$ approval rating.
For comparison of the new labels with the original \texttt{EDOS} and \texttt{DOM} labels, we recruited people based in the same region as the original annotators, the United Kingdom. 
We further collected demographic information and responses to questions from three surveys designed to measure the attitudes of workers.




\paragraph{Measurement of Attitudes}

To measure the annotators' attitudes, we used survey questions from two verified scales widely used in social psychology to measure the constructs described in \autoref{sec:background}: the \emph{Very Short Authoritarianism} (VSA)~\citep{bizumic2018investigating} and \emph{Short Social Dominance Orientation} (SSDO)~\citep{pratto2013social} measuring RWA and SDO, respectively. We also collected responses to the five questions of the \emph{Brief Hostile Neosexism Scale} (BHNS) to measure HN \citep{chulvi-etal-2023-social}.
As shown in Figure \ref{fig:boxes} of \autoref{sec:survey_responses}, overall attitudes show tendencies towards social dominance and neosexism, but not towards authoritarianism, although attitudes on all scales vary considerably among the annotator pool. See \autoref{sec:background} and \autoref{app:scales} for more details.\looseness=-1 


\paragraph{Data Labelling}
\citet{chulvi-etal-2023-social} have shown that the responses of around 12 annotators per item are sufficient to capture levels of disagreement for a similar sexist language labelling task.
We collect up to 23 labels per item to enable investigation in this task.
 We provide annotators with the relevant parts of the original annotator instructions and guidelines from \citet{kirk-etal-2023-semeval} and \citet{guest-etal-2021-expert}.
Instructions are provided in \autoref{sec:intructions}.

\paragraph{Intra-Annotator Agreement}
We measure agreement between our recruited annotators as well as between the aggregated labels, decided by majority vote, and the original \texttt{EDOS} and \texttt{DOM} labels.
We report raw percentage agreement and Krippendorf's $\alpha$, which measures agreement between two or more raters and can handle missing values~\citep{gwet2014handbook}.

\vspace{-5pt}
\begin{table}[ht!]
    \centering
    \small
    \begin{tabular}{l|p{1.75cm}|l}
         \textbf{Crowd workers} & \multicolumn{2}{c}{\textbf{Majority vote \emph{v} Original labels}} \\
         $\alpha$ & $\alpha$ & \% \\
         \hline 
         $0.02$ & $0.25$ & $70.8$  
    \end{tabular}
    \caption{Reliability measured by inter-annotator agreement (Krippendorf's $\alpha$ and percentage agreement ($\%$)).} 
    \label{tab:iaa}
\end{table}
\vspace{-6pt}

As shown in Table \ref{tab:iaa}, agreement between the crowd-sourced annotators is low at only $\alpha=0.02$, although aggregated labels are more similar to the original labels (also produced by majority vote).



\section{Statistical Analysis}
\label{sec:stats}

Our hypothesis is two-tailed and exploratory in nature: whether gender, SSDO, BHNS, or VSA scores are predictive of annotator behaviour in labelling items as sexist/misogynist. 

\paragraph{Experimental Design}


Since we have multiple annotations per annotator, we employ a mixed effects regression model \citep{raudenbush1994random}. 
Our dependent variable is the binary label given by each annotator, while our predictors are gender (\emph{male}/\emph{female}), SSDO and BHNS scores (both aggregated into \emph{High}, \emph{Moderate} and \emph{Low}), and VSA scores (aggregated into a five-point scale from \emph{very low} to \emph{very high}). 
Our model includes by-annotator random intercepts, as most individuals annotated multiple items, while we reject one participant who only provided a single annotation. 
All categorical variables are dummy-coded (see \autoref{sec: dummy_code} for details). 

To evaluate possible effects of pairwise comparisons, we employ a Benjamini-Hochberg correction \citep{benjamini1995controlling} due to its specific focus on false discovery rates in study designs with independent statistics, and smaller sample sizes \citep{benjamini2000adaptive,thissen2002quick}, such as our study.

\paragraph{Results}

\begin{table}[!ht]
\centering
\scriptsize

\sisetup{mode = match, tight-spacing=true,table-number-alignment=center,separate-uncertainty=true,table-align-uncertainty=true,detect-weight=true}
\renewcommand{\arraystretch}{1.3}
\begin{tabularx}{\linewidth}{@{}X S[table-format=-1.2] *{2}{S[table-format=1.2]} S[table-format=-1.2] S[table-format=1.3] S[table-format=1.3] @{}}
    & {Estimate} 
    & {SE} 
    & {$z$} 
    & {$p$} 
    & {$p$\textsubscript{adj.}} 
    \\
    \midrule
    Intercept & 0.85 & 0.58 & 1.48 &  0.140 & 0.251\\
    Gender-Male & 1.16 & 0.54 & 2.14 &  \bfseries 0.032 & 0.073\\
    VSA - Low & -0.31 & 1.11 & -0.28 &  0.783 & 0.951\\ 
    VSA - High & 2.61 & 0.96 & 2.73 &  \bfseries 0.006 & \bfseries 0.039\\
    VSA - V. High & 0.73 & 1.21 & 0.60 &  0.548 & 0.822\\
    SSDO - Low & 0.06 & 0.82 & 0.08 &  0.937 & 0.951\\
    SSDO - High & -1.71 & 0.69 & -2.48 &  \bfseries 0.013 & \bfseries 0.039\\
    BHNS - Low & 0.07 & 1.09 & 0.06 &  0.952 & 0.951\\
    BHNS - High & -1.54 & 0.60 & -2.59 &  \bfseries 0.010 & \bfseries 0.039\\   
    
\end{tabularx}
\caption{Regression model evaluative outcomes. P\textsubscript{val.} refers to significance of initial findings; P\textsubscript{adj.} refers to the adjusted P\textsubscript{val.} after a Benjamini \& Hochberg correction.}
\label{tab:stat_results}
\end{table}

The results of our regression analysis are shown in \autoref{tab:stat_results}. 
Our post-hoc correction resulted in our initially significant result on the effects of gender being rejected. Nevertheless, we report a significant positive effect of the VSA-High condition on rating items as sexist (estimate = 2.61, SE  = 0.96, $z$ = 2.73, $p$= 0.006, $p$\textsubscript{adj.} = 0.039). We further report a significant negative effect of the SSDO-High condition in annotating items as sexist (estimate = -1.71, SE  = 0.69, $z$ = -2.48, $p$= 0.013, $p$\textsubscript{adj.} = 0.039). Finally, we report a strong negative effect of the BHNS-High condition on annotating items as sexist (estimate = -1.54, SE =  0.60, $z$ = -2.59, $p$=  0.010, $p$\textsubscript{adj.} = 0.039).

\paragraph{Discussion}

Our findings echo prior work showing that demographics 
do not always influence annotation behaviour
\citep{beck-etal-2024-sensitivity,biester-etal-2022-analyzing,orlikowski-etal-2023-ecological}. 
However, our 
results suggest a directional effect of the annotators' attitudes: higher VSA scores predict hypersensitivity in annotating sexism, indicating a positive propensity to label items as sexist. Conversely, higher scores along
the SSDO and BHNS scales predict lower levels of annotations of sexism. 

These findings are particularly interesting if placed within the context of the constructs that the scales themselves measure. Since RWA has been shown to be associated with benevolent sexism \citep{de2022understanding}, this could explain why annotators with higher VSA scores demonstrate a higher propensity to label items as sexist. 
We should note that our results show a significant effect only in the VSA-High condition, not the VSA-Very High condition, despite the trend of the effects being in the same direction. This suggests that while there is a significant difference in annotation behaviour between the high and moderate VSA groups, this effect does not extend consistently to the very high VSA group. We will address this inconsistency in our future work and conduct further analysis to explore potential factors, such as uneven sample sizes or demographic influences, that may have contributed to this unexpected result.

In contrast, with SDO and HN being linked to hostile sexism \citep{la2020social, chulvi-etal-2023-social}, the effect found by their respective scale might explain annotator leniency towards sexist items, aligning with prior work by \citet{sap-etal-2022-annotators}, i.e., on annotators with racist beliefs showing similar leniency towards racist language. 
This suggests that different dimensions of authoritarianism, social dominance, and neosexism can influence the nature of bias in annotations.

\section{Classification Experiments}
\label{sec:experiments}
    



We benchmark the new dataset to explore i) whether a broader label set provides richer information and ii) how varying conceptualisations of GBV,
annotator demographics and attitude information affect model performance in identifying GBV text online.
To achieve this, we conduct two tasks to predict (a) majority labels per text, and (b) individual annotator labels with diverse label texts.

\paragraph{Dataset}

We use our re-annotated dataset introduced in \autoref{sec:datasets} to predict the majority label per text. Our dataset contains 1600 instances,
including 1200 instances from \texttt{EDOS} subset and 400 from \texttt{DOM}. 
We also augment it with individual annotator labels, obtaining 6,000 and 23,000 instances for the fine-tuning and test sets respectively.
Three different label texts are used: ``GBV'' as the aggregated label in our re-annotated dataset, ``Sexist'' from \texttt{EDOS}, and ``Misogynistic'' from \texttt{DOM} (see \autoref{tab:classes}). 

\paragraph{Prompt Design}

We experiment with five different prompt templates for our detection task.

\textbf{(1) Label prompt}: a simple prompt structure to give the label based on the text alone. 
The template is ``\texttt{<text>} This text is classified as''.

\textbf{(2) Task description (task)}: starts with an instruction describing the detection task, followed by the text to be classified. 
The instruction for the task description is ``Classify the following text from a social media platform. It might contain a form of \texttt{<label>}. Output \texttt{<label>} if it contains \texttt{<label>}, or not \texttt{<label>} if not.'', and the template is ``\mbox{\texttt{<task description>}} Text: \texttt{<text>} \texttt{<choices>} Answer:''.

\textbf{(3) Few-shot demonstrations (demos)}: incorporates the task description and adds two examples (demonstrations) of texts with their corresponding labels before the text is classified.
The template is ``\texttt{<task description> <demonstration>} Text: \texttt{<text>} \texttt{<choices>} Answer:''.

\textbf{(4) Annotator descriptions (anno)}: combines the task description with a description of the annotator's demographic and attitude information before the text. The annotator description can be either a full description of all questions and answers from questionnaires (full) or a brief description of each scale (short), plus its corresponding range based on the compound score\footnote{A compound score is a unified measure derived by aggregating individual responses to multiple questions in a questionnaire, enabling quantification and comparison. More details are provided in \autoref{app:scales}.} for the annotator.
The template is ``\texttt{<task description> <annotator description>} Text: \texttt{<text>} \texttt{<choices>} Answer:''.

\textbf{(5) Combined prompt}: integrates the task description, few-shot demonstrations, and annotator description before the target text. For each demonstration, we add two annotators' descriptions and labels.
The template is ``\texttt{<task description> <demonstration> <annotator description>} Text: \texttt{<text>} \texttt{<choices>} Answer:''.

We use the answer format from 
\citet{eval-harness}. 
\texttt{<choices>} is described as ``Choices: A. GBV or B. Not GBV.'' for ``GBV'' label text, and corresponding changes are made for ``Sexist'' and ``Misogynistic'' labels (see \autoref{app:input} for more details). 

\paragraph{Models} We conduct two sets of experiments, namely ICL and fine-tuning.\footnote{We use low-rank adaptation (LoRA) \citep{hu2022lora} for all models to reduce the number of trainable parameters.}

\textbf{(1) \texttt{RoBERTa}$_{base}$, \texttt{RoBERTa}$_{hate}$}: we perform ICL only to smaller encoder-only pre-trained LMs. The latter has been pre-trained on toxic language datasets making it a good candidate for the GBV classification task.

\textbf{(2) \texttt{FLAN-T5}} : 
we perform both ICL and fine-tuning experiments with a (much larger) encoder-decoder instruction fine-tuned LLM. 
Fine-tuning enhances the model's reliability, while the subjectivity of GBV classification makes it difficult for an ICL model to capture the relationship between annotator attitudes and behaviours to labels with only few-shot demonstrations.

\textbf{(3) \texttt{LLaMA 2 7B}, \texttt{LLaMA 3 8B}, \texttt{Mistral 7B}}: 
We \textit{fine-tune} only the base version of three decoder-only LLMs. \texttt{LLaMA 2} 
and \texttt{LLaMA 3} 
have been shown to adapt to new tasks with relatively few instructions \citep{milios-etal-2023-context}, making them ideal for our low-resource setting (600 training instances), while \texttt{Mistral} 
exhibits significant performance, especially on text classification tasks. 

\paragraph{Experimental Design}

To predict the majority labels for the GBV detection task, we apply the ICL experiment directly to the test set with 1000 instances, only using the ``Label Prompt'' template for inference. 
For the individual annotator label prediction task, we conduct three ICL experiments on \texttt{FLAN-T5} with three different label texts used for the whole set, and another ICL for original labels from two subsets respectively (predicting instances from \texttt{EDOS} using ``Sexist'' and those from \texttt{DOM} using ``Misogynistic'').
Then fine-tuning experiments use only ``GBV'' label with all four LLMs.
Both experiments test our augmented re-annotated dataset and its two subsets and utilise different prompts (i.e. prompt templates 2-5) under zero-shot and few-shot scenarios, to further investigate the influence of individual annotator's behaviours.
Given skewed label distribution, we report the macro F1 score; for hyperparameter settings, see \autoref{app:experiment}.


\paragraph{Results and Analysis}

Table \ref{tab:experiment-icl-majority} shows classification results on majority labels via ICL. All three models outperform the majority-class baseline on both sets of annotations. However, \texttt{RoBERTa}$_{base}$ does so only marginally. Results from \texttt{RoBERTa}$_{hate}$ underline the strength of models tailored for a specific task, such as GBV detection here. 
\texttt{FLAN-T5} outperforms all models, showcasing its superior capability when lacking annotated datasets.


\begin{table}[ht!]
    \centering
    \small
    \begin{tabular}{l|c|c}
         \textbf{Model} & \textbf{Original Annotation} & \textbf{Re-annotation} \\
        \toprule
        Majority class & $44.54$ & $40.72$   \\
        \texttt{RoBERTa}$_{base}$ & $45.77$ & $42.11$ \\
        \texttt{RoBERTa}$_{hate}$ & $52.05$ & $48.65$  \\
        \texttt{FLAN-T5} & $\mathbf{61.40}$ & $\mathbf{57.55}$ 
    \end{tabular}
    \caption{Results of predicting majority labels via in-context learning for the GBV detection task.}
    \label{tab:experiment-icl-majority}
\end{table}

Table \ref{tab:experiment-icl-single-flant5} presents ICL results for \texttt{FLAN-T5} using different label texts and input prompts on the augmented dataset.
Among the four label settings, 
predicting ``Sexist'' on our full dataset consistently outperforms the others based on re-annotated labels, achieving the highest score of 65.62. 
Using the \texttt{DOM} label ``Misogynistic'' also performs better with short annotator descriptions. 
Regarding the datasets, 
better performance is generally achieved on the benchmark
when compared to two subsets.
Besides, adding annotator descriptions or demonstrations usually leads to superior performance,
while combining both annotator information and demonstrations does not always enhance performance, highlighting the importance of proper prompt design.

\begin{table*}[ht!]
    \centering
    \small
    \begin{tabular}{l|c|c|c|c|c|c|c|c|}
         \textbf{Model: \texttt{FLAN-T5}} & \multicolumn{3}{c|}{\textbf{Original}}  & \multicolumn{3}{c|}{\textbf{GBV}} & \textbf{Sexist} & \textbf{Misogynistic}\\    
         & \texttt{All} & \texttt{EDOS} & \texttt{DOM} & \texttt{All} & \texttt{EDOS} & \texttt{DOM} & \texttt{All}  & \texttt{All}\\
         \toprule
        Majority class (single) & $36.14$ & $35.71$ & $37.41$ & $36.14$ & $35.71$ & $37.41$  & $36.14$ & $36.14$\\
        task &$60.93$& $\mathbf{64.87}$ &  $56.95$& $60.29$ & $60.53$ & $56.81$ & $65.25$ & $61.12$\\
        +demos &$59.75$&$64.24$&$54.25$& $\mathbf{62.60}$ & $\mathbf{62.92}$ & $\mathbf{58.49}$ & $63.11$ & $59.67$ \\
        +anno (short) & $\mathbf{64.68}$ &$64.28$&$\mathbf{62.15}$& $61.13$ & $61.35$ & $57.05$ & $\mathbf{65.62}$ & $\mathbf{64.69}$ \\
        +anno (short)+demos &$62.42$&$63.72$&$59.32$& $59.91$ & $60.50$ & $54.80$ & $63.43$ &  $62.32$  \\
        +anno (full)  &$59.03$&$62.21$&$53.97$& $61.22$ &  $61.38$ & $57.65$ & $62.23$ & $58.20$ \\
        +anno (full)+demos  &$59.03$&$62.21$&$53.97$& $61.22$ & $61.38$ & $57.65$ & $62.23$ &  $58.20$ \\ 
    \end{tabular}
    \caption{Results of in-context learning on our re-annotated dataset using \texttt{FLAN-T5} with different label texts: (i) ``Original'' uses the original labels, namely ``Sexist'' for \texttt{EDOS} subset and ``Misogynistic'' for \texttt{DOM} subset, (ii) ``GBV'' as the aggregated label for both subsets, (iii) ``Sexist'' and (iv) ``Misogynistic''  for both subsets. Six different input prompts are evaluated among three label texts. Best results are shown in bold by column.}
    \label{tab:experiment-icl-single-flant5}
\end{table*}

\begin{table*}[ht!]
    \centering
    \small
    \begin{tabular}{l|c|c|c|c|}
         \textbf{New Label - GBV} (maj. $36.14$) & \textbf{\texttt{FLAN-T5}}  & \textbf{\texttt{LLaMA 2}} &  \textbf{\texttt{LLaMA 3}}  & \textbf{\texttt{Mistral}} \\ \toprule
        task &$63.78\pm1.84$  &  $51.87\pm1.76$ &  $50.32\pm2.75$ & $59.20\pm2.10$ \\
        +demos & $65.12\pm1.66$ &$49.40\pm1.79$  & $\mathbf{52.12\pm1.05}$ & $41.07\pm1.18$ \\
        +anno (short) &  $\mathbf{65.79\pm1.89}$ & $51.17\pm1.58$ &  $43.39\pm1.47$ &  $52.56\pm1.79$\\
        +anno (short)+demos & $64.95\pm1.03$ &  $41.16\pm1.06$ & $50.40\pm0.53$  & $\mathbf{67.40\pm1.55}$\\
        +anno (full) & $64.50\pm1.17$ & $\mathbf{53.21\pm0.12}$ & $51.70\pm2.51$ & $40.96\pm2.97$ \\
        +anno (full)+demos & $62.23\pm0.54$ & $50.02\pm1.10$ &  $43.31\pm0.29$& $54.15\pm0.46$\\
    \end{tabular}
    \caption{Results of fine-tuning LLMs on individual annotator labels using different input prompts for the GBV detection task. F1 score for the majority class (maj.) is $36.14$. Best results are displayed in bold by column.}
    \label{tab:experiment-ft-single}
\end{table*}

Table \ref{tab:experiment-ft-single} shows results of fine-tuning experiments 
on individual annotator labels using various input prompts for the GBV detection task. 
Unsurprisingly, \texttt{FLAN-T5} outperforms the ICL variant (Table \ref{tab:experiment-icl-single-flant5}). 
Fine-tuning is used to improve the model's reliability for annotators. 
Given that GBV classification is a subjective task, it can be challenging for an ICL model to capture the relationship between nuanced annotator metadata (attitudes) and their behaviours (labels) with just few-shot demonstrations or definitions.
See Appendix \ref{app:experiment-result} for further results. 
Among the four LLMs, \texttt{FLAN-T5} outperforms \texttt{LLaMA 2}, \texttt{LLaMA 3}, and \texttt{Mistral},
showing significant performances with 
short annotator description for the new label.
Besides, the effectiveness of input prompts varies. Adding full annotator descriptions generally provides better results, 
particularly for \texttt{LLaMA 2} and \texttt{LLaMA 3}, and inputs with short annotator descriptions also improve the results. 
The combined prompts with short annotator information and demonstrations, especially for \texttt{Mistral}, show great improvements.
These results suggest that more comprehensive information generally enhances model performance, but the effectiveness can vary by model.

\paragraph{Discussion}




% ...
We explore the GBV classification pipeline with an emphasis on the role of diverse annotator demographics and attitudes.
Our classification experiments in  \autoref{sec:experiments} reveal that detailed prompts, enriched with annotator bias, significantly improve LLM adaptability in identifying GBV content, even in low-resource scenarios. 
This highlights the importance of well-crafted input prompts in enhancing the model's ability to accurately interpret and respond to complex social phenomena involving variably interpretable elements.

However, a contrasting result is presented when we explore the use of annotator information with demonstrations in ICL and fine-tuning experiments.
We observe a decrease in model performance when prompts are overloaded with additional information.
This indicates that the quality of input prompt can influence the model's efficiency and additional information may hinder rather than help performance. 
Statistical findings in \autoref{sec:stats} reveal that annotators' personal biases affect their labelling tendencies. Pre-trained models may lack perspectivist information, and adding a few demonstrations is insufficient for the model to learn these biases, potentially causing confusion. 
Besides, the \texttt{Flan-T5} model is trained on samples with a maximum length of 1024 tokens. As it uses relative positional encoding, it is possible to train the model with far more tokens, but training with samples with a greater length could hurt model performance \citep{chung2022scaling}.
It is also possibly because the models get negative biases from the particular demonstrations used,  affecting their ability to accurately interpret GBV-related content.

In our analysis of the \texttt{FLAN-T5} model's performance using different label texts, we found that ``Sexist'' leads to the best results, followed by ``Misogynistic'', with ``GBV'' performing the poorest, which might be attributed to several factors. 
These three labels have different conceptualizations of sexism.
``Sexist'' is relatively straightforward and less ambiguous,
whereas ``Misogynistic'' is more specific, introducing stronger connotations to complicate classification.
``GBV'' is broader and more complex, covering violence and discrimination in various contexts and forms. It may increase cognitive load and interpretation difference.
Additionally, the original labels from \texttt{EDOS} and \texttt{DOM} might align better with ``Sexist'' and ``Misogynistic'' classifications, respectively, compared to ``GBV''.
Furthermore, ``Sexist'' and ``Misogynistic'' could be more understandable for models based on previous pre-trained resources. 
Therefore, it is essential to develop a clearer and more precise definition and taxonomy of gender-based violence to enhance the model's learning and classification abilities.


Furthermore, our analysis of various experiments on the newly annotated labels indicates a generally poorer performance compared to the original labels, which could be attributed to increased complexity, annotator bias, and class imbalance in the new annotation set.
This emphasises the challenge of adapting automated classification models to variably interpretable and evolving social issues such as GBV detection, where language complexity and human perspectives intersect.\looseness=-1 


\section{Conclusion}

We have revisited the annotation and classification tasks for online GBV with a particular focus on the underlying attitudes of a broad range of annotators.\looseness=-1

Through the re-annotation of two 
datasets, we collect demographic and attitudinal information about crowd-sourced annotators using validated surveys from Social Psychology, and incorporate a diverse range of annotator perspectives, exploring the relationship between these factors and the labels they provide.
We find that annotators with stronger right-wing authoritarian traits show a higher propensity to label items as sexist, whereas those with more socially dominant and neosexist attitudes do the opposite. This suggests that people exhibiting right-wing authoritarian characteristics may be less attuned to subtle gender-related discourse. \looseness=-1

We then conduct classification experiments on both aggregated and individual annotator labels using various prompting strategies and LLMs.
Our findings indicate that models are biased by annotators' attitudes. While incorporating annotator information can enhance model capacity, but adding excessive information can be detrimental.
This highlights the challenges of the increased complexity and imbalance of incorporating broader, \emph{perspectivist} label sets, which adversely affect performance on all our experiments.


\section*{Limitations} \label{sec:limitations}

This study 
concentrates on sexism and misogyny from the scope of the \texttt{EDOS} and \texttt{DOM} datasets. Future research directions require a broader GBV framework that captures the full spectrum of GBV-related issues and more inclusive dataset standards.

We recruit annotators only from MTurk, who might provide unreliable data on personal information~\citep{huang-etal-2023-incorporating} 
or sensitive topics such as GBV, raising potential data quality issues.

Due to the exploratory and experimental nature of this work, the statistical annotation analysis suffers from a number of important limitations,
namely the relatively small sample size and unequal amount of annotations per annotator.
Future studies should aim for a fixed number of annotations per annotator, while a larger sample size would lead to more generalisable results.
There is also the need for further clarification and nuanced interpretation of the statistical results, especially in the context of very high VSA scores.

Lastly, our classification experiments and analysis are limited to open-source LLMs such as \texttt{Flan-T5}, \texttt{LLaMA 2}, \texttt{LLaMA 3}, and \texttt{Mistral}. The exclusion of other LLMs, such as the GPT family, limits the reproducibility and breadth of our work. The training resources for LLMs we used are in limited languages, which may not accurately capture GBV in multilingual contexts. It indicates the need for future work to explore a wider range of LLMs to model GBV online.\looseness=-1


Although we account for annotator attitudes in our study to explore their impact on labelling, accurately measuring the effect of these attitudes remains a challenge. While this approach can provide insights into potential biases, it does not fully resolve the issue of maintaining objectivity in tasks like sexism and misogyny detection. Ensuring that annotations are both accurate and objective still requires a careful selection of annotators with relevant expertise and experience.

\section*{Ethical Considerations}

This study was approved by the Institutional Review Board (IRB) of 
the School of Mathematical and Computer Sciences at Heriot-Watt University,\footnote{Project identification code is 5536.} which reviewed our annotation methodologies and protocols to ensure compliance with ethical standards.\looseness=-1

As annotators were exposed to potentially upsetting language, we took the following mitigation measures: 

\begin{itemize}
    \item Participants were warned about the content (1) before accepting the task on the recruitment platform, (2) in the Information Sheet provided at the start of the task, and (3) in the Consent Form where they acknowledged the potential risks.
    \item Participants were required to give their consent to participation.
    \item They were able to leave the study at any time on the understanding that they would be paid for any completed work.
    \item The task was kept short (all participants completed each round in under 30 minutes) to avoid lengthy exposure to upsetting material.
\end{itemize}

Following the advice of \citet{shmueli-etal-2021-beyond} we paid participants at a rate that was above both  Prolific's current recommendation of at least \textsterling9.00 GBP/\$12.00 USD\footnote{\href{https://www.prolific.co/blog/how-much-should-you-pay-research-participants}{\texttt{https://www.prolific.co/blog/how-much- should-you-pay-research-participants}}} and the Living Wage in our jurisdiction, which is considerably higher. 

We follow the advice of \citet{kirk-etal-2022-handling} on presenting harmful text both to annotators and to the readers of this document.

Due to the size of our annotation pool, for this study, analysis of annotators' demographic characteristics was limited to individual features.
We recognise that responses to GBV are influenced by complex intersectional identities that we have been unable to capture here, but which will be the focus of future data collection and analysis. 





\section*{Acknowledgements}
We would like to thank the anonymous reviewers for all their valuable comments and suggestions.
Aiqi Jiang, Tanvi Dinkar, Gavin Abercrombie, and Ioannis Konstas were supported by the EPSRC project ``Equally Safe Online'' (EP/W025493/1).
This work was supported by the Heriot-Watt University high-performance computing facility (DMOG) and the Edinburgh International Data Facility (EIDF).
\bibliography{anthology,anthology_p2,ref}
\bibliographystyle{acl_natbib}

\appendix

\section{Data Statement}
\label{app:data_statement}

We provide a data statement to summarise the main features of the selected datasets, as recommended by \citet{mcmillan-major-etal-2023-data}.

\paragraph{Curation rationale} Textual data is from the test set of \texttt{EDOS} and \texttt{DOM}, selected for the reasons highlighted in \autoref{sec:datasets}.
For further details of the original data collection processes, see \citet{kirk-etal-2023-semeval} and \citet{guest-etal-2021-expert}.

\paragraph{Language variety:} \texttt{en}. English, as written in comments on internet forums on the Gab and Reddit platforms.

\paragraph{Author demographics:} According to \citet{kirk-etal-2023-semeval}, post authors are `are likely male, western and right-leaning, and hold extreme or far-right views about women, gender issues and feminism'.
No information is available regarding authors of \texttt{DOM} texts. 


\paragraph{Annotator demographics:}
\begin{itemize}
    \item Age: $24-56$, $m=35.8$, $s=7.9$
    \item Gender: Female: $19$ ($44.2\%$); Male: $24$ ($55.8\%$). 
    \item Ethnicity: White: $35$ ($81.4\%$); Asian: $6$ ($14.0\%$); Black: $1$ ($2.3\%$); Other: $1$, ($2.3\%$).
    \item Sexual orientation: Heterosexual: $23$ ($53.5\%$); Bisexual: $18$ ($41.9\%$); Don't know: $1$ ($2.3\%$); Prefer not to say: $1$ ($2.3\%$). 
    \item Political orientation: Left-wing/liberal: $7$ ($16.3\%$); Centre $16$ ($37.2\%$); Right-wing/conservative $16$ ($37.2.\%$); None/prefer not to say: $4$ ($9.3\%$).
    \item Training in relevant disciplines: Unknown
\end{itemize}

\paragraph{Text production situation:}
\begin{itemize}
    \item Time and place: August 2016 to October 2018; Gab and Reddit.
    \item Modality: Text.
    \item Intended audience: Internet forum users.
\end{itemize}

\paragraph{Text characteristics} 
The posts were taken from forums known to attract misogynistic rhetoric: Gab, an extreme-right leaning forum and subreddits labelled as `Incels', `Men Going Their Own Way', `Men’s Rights
Activists', and `Pick Up Artists'.
\citet{kirk-etal-2023-semeval} also provides a full data statement.

\section{Measuring Social Attitudes} \label{app:scales} 

The VSA scale \citep{bizumic2018investigating} is a modified version of the original RWA \citet{altemeyer1983right}, which reduced the original 30-item questionnaire into 6 items, while the SSDO scale is a modified version of the original SDO developed by \citet{pratto1994social}, which reduced the original 16-item scale into 4 items. 
Both scales have been verified towards both internal and external validity while ensuring that all elements of the original subscales are adequately captured \citep{altemeyer1983right, pratto1994social}.

Furthermore, both the VSA and the SSDO scales have been verified through a variety of cultures and contexts \citep{aichholzer2021refining, pratto2013social, mcbride2021monitoring, azevedo2019neoliberal, tonkovic2021believes}. 
Each participant answered through the full battery of questions present in each questionnaire, as removing a subsection of items can invalidate the questionnaire responses \citep{jebb2021review}. 
The full lists of items are presented below.

\subsection{Very Short Authoritarianism Scale (VSA)} \label{app:vsa_scale}

The scale reporting was based on a 9-point Likert scale, ranging from Very strongly disagree to Very strongly agree. The scale is consist of sub-dimensions, namely Conservativism, Authoritarianism, Traditionalism, Authoritarian Aggression and Authoritarian Submission. Letter R indicates that the item is reverse scored.

\begin{itemize}
    \item It’s great that many young people today are prepared to defy authority. (Conservatism or Authoritarian Submission)- (\textbf{R})
    \item What our country needs most is discipline, with everyone following our leaders in unity (Conservatism or Authoritarian Submission)
    \item God’s laws about abortion, pornography, and marriage must be strictly followed before it is too late. (Traditionalism or Conventionalism)
    \item There is nothing wrong with premarital sexual intercourse. (Traditionalism or Conventionalism) (\textbf{R})
    \item Our society does NOT need tougher Government and stricter Laws. (Authoritarianism or Authoritarian Aggression) (\textbf{R})
    \item The facts on crime and the recent public disorders show we have to crack down harder on troublemakers, if we are going to preserve law and order. (Authoritarianism or Authoritarian Aggression) 
\end{itemize}

With 6 questions and a 9-point scale, the score range for each question is 1 to 9.
The compound score is calculated by summing the scores across these 6 items and adjusting for reverse scoring where applicable. This total score categorises respondents into five levels of RWA: very low, low, moderate, high, and very high, ranging from 6 to 54.

\begin{itemize}
    \item Very low right wing authoritarianism: 6-15
    \item Low right wing authoritarianism: 16-25
    \item Moderate right wing authoritarianism: 26-35
    \item High right wing authoritarianism: 36-45
    \item Very high right wing authoritarianism: 46-54
\end{itemize}

While previous studies have used the scale with three breakpoints (low, moderate, and high), there is evidence to suggest that the moderate range might be concealing effects between the combinations of sub-dimensions that form the original RWA scale, and thus the VSA \citep{funke2005dimensionality}. To address this, we follow the original guidelines set by the creator of the RWA scale, from very low to very high \citep{altemeyer1996authoritarian,smith1999authoritarian}, allowing for a finer distinction between the data that stays true to methodologies previously used to study the construct.

\subsection{Short Social Dominance Orientation Scale (SSDO)} \label{app:ssdo_scale}

The scale reporting was based on a 7-point Likert scale, ranging from Strongly disagree to Strongly agree. All emphasis in text was also present in the original SSDO scale. For items 2 and 4, higher numeric values indicate a higher level of SSDO and are weighted higher.

\begin{itemize}
    \item In setting priorities, we must consider all \textit{societal} groups. 
    \item We should not push for equality of \textit{societal} groups. 
    \item The equality of \textit{societal} groups should be our goal. 
    \item Superior \textit{societal} groups should dominate inferior groups. 
\end{itemize}

With 4 questions and a 7-point scale, the score range for each question is 1 to 7.
The compound score is calculated by summing the scores across these 4 items and adjusting for reverse scoring where applicable. This total score categorises respondents into three levels of SDO: low, moderate, and high, ranging from 4 to 28. 

\begin{itemize}
    \item Low social dominance orientation: 4-10
    \item Moderate social dominance orientation: 11-17
    \item High social dominance orientation: 18-28
\end{itemize}

\subsection{Brief Hostile Neosexism Scale (BHNS)} \label{app:neosexism_scale}

\citet{chulvi-etal-2023-social}'s scale is based on a 7-point Likert scale, ranging from \emph{Strongly disagree} to \emph{Strongly agree}. 
All emphasis in text was also present in the original neosexism scale. 
For all 6 items, higher numeric values indicate a higher level of hostile neosexism and are weighted higher.

\begin{itemize}
    \item Some of the demands of the feminist movement seem to me to be a bit exaggerated. 
    \item I sometimes feel that our society pays too much attention to the rights of certain minorities. 
    \item In the name of equality, many women try to gain certain privileges.
    \item Many women interpret innocent comments and actions as sexist.
    \item Women are easily offended.
    \item Women exaggerate the problems they suffer because they are women.
\end{itemize}

With 6 questions and a 7-point scale, the score range for each question is 1 to 7.
The compound score is calculated by summing the scores across these 6 items and adjusting for reverse scoring where applicable. This total score categorises respondents into three levels of hostile neosexism: low, moderate, and high,  ranging from 6 to 42. 

\begin{itemize}
    \item Low hostile neosexism: 6-14
    \item Moderate hostile neosexism: 15-28
    \item High hostile neosexism: 29-42
\end{itemize}


\subsection{Survey Responses} \label{sec:survey_responses}

\begin{figure}[ht!]
    \centering
    \includegraphics[width=\columnwidth]{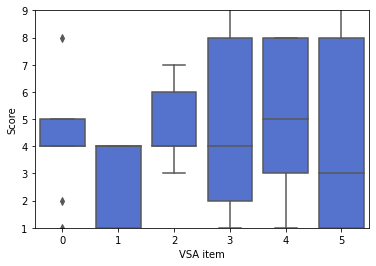}
    \includegraphics[width=\columnwidth]{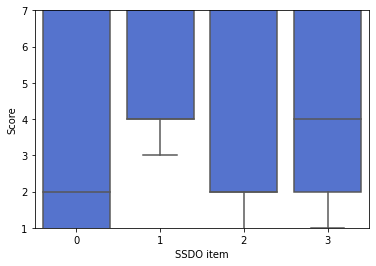}
    \includegraphics[width=\columnwidth]{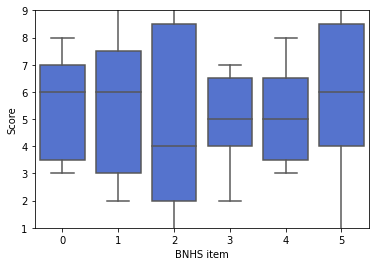}
    \caption{Responses to the six VSA, four SSDO, and six BHNS survey items on $[1-9]$, $[1-7]$, and $[1-7]$ scales, respectively.}
    \label{fig:boxes}
\end{figure}

Annotator responses to the survey questions are presented in \autoref{fig:boxes}.

We find that for VSA, the annotators tend towards the centre of the scale ($m=4.78, s=3.35$), 
while for SSDO, they are towards the more dominant end of the scale on average ($m=6.14, s=5.19$), and for BHNS, they tend towards hostile neosexist attitudes ($m=4.78, s=3.35$), as shown in Figure \ref{fig:boxes}. 
There is, however, substantial variance on all three scales.
To sum up, overall attitudes show tendencies towards social dominance and neosexism, but not towards authoritarianism, and attitudes on all scales vary considerably among the annotator pool.

\section{Statistical Testing} \label{sec: dummy_code}

All statistical analyses were done in R \citep{R_core}, with the packages: 1.3.1 tidyverse \cite{tidyverse}, lme4 \citep{lme4}, simR 1.0.7 \citep{simR}.

The reference groups for our dummy coding were as follows:

\begin{itemize}
    \item Gender: Women
    \item VSA: Moderate scores
    \item SSDO: Moderate scores
    \item BHNS: Moderate scores
\end{itemize}

We chose the moderate scale for all of our scales as a baseline as previous research has shown that the effects of social attitudes tend to influence behaviour when scoring alongside the edges of a scale (either low or high) \citet{davani-etal-2023-disentangling, hettiachchi2023crowd}.

\section{Annotation Instructions} \label{sec:intructions}


We provide an annotation instruction and the annotators must read and accept it before they start annotating texts. We give a content warning before the instruction: This research exposes participants to offensive language which may cause mental or physical stress to the reader. Please consider this before participating, you are under no obligation to take part and if you choose not to we thank you for considering taking part. Please do remember that you can withdraw from participating at any time. 

Our full annotation instruction is described below:

\paragraph{Sexism}
This task defines sexism as: "Any abuse or negative sentiment that is directed towards women based on their gender, or on the combination of their gender with one or more other identity attributes (e.g. Black women, Muslim women, Trans women)"

An entry must be labelled Sexist if it meets both of the following conditions:

\begin{enumerate}
    \item The entry refers to a woman, a group of women, women in general, or to supporters of feminism. For this task, "woman" refers to any person who identifies as a woman, irrespective of gender assigned at birth (i.e., include transgender women within this definition). In addition, explicitly threatening or inciting harm against individual women must also be included.
    \item The entry expresses negative sentiment against its target on the basis of gender: for instance, it is derogatory, demonising, insulting, threatening, violent or prejudicial.
\end{enumerate}

Your task is to label the entry rather than the speaker. Even in cases where the speaker could be sexist, please carefully consider whether the statement itself meets the above criteria.

\textbf{Notes on quotes and jokes:} Entries which make a quote (indicated by "") without any further comments should be taken at face value. If an entry contains a joke, please carefully consider its intention. If a joke meets the above criteria, it should be labelled as Sexist and put into the corresponding secondary category, even in cases where the tone is light-hearted or positive.

\paragraph{Not Sexist}

For each Not Sexist entry, you need to decide whether it contains abuse directed at another protected characteristic (i.e., a fundamental aspect of a person's identity) besides gender. Examples of other protected characteristics include, but are not limited to: race, ethnicity, immigration status, religion, age, sexuality, and disability status. If it does, write out the target of the abuse.

\textbf{Common types of confusing Not Sexist content:}
Some entries that you should label Not Sexist may easily be confused with Sexist content. Please review the following examples, all of which should be labelled Not Sexist:

\begin{enumerate}
    \item Entries which contain vulgar, inappropriate or offensive language, but do not specifically target women, e.g.,

    \begin{itemize}
        \item "We're here at the bar, now suck my penis"
        \item "Hahahaa you silly dickhead"
    \end{itemize}

    \item Entries that direct abuse against individuals without a gender-based attack, e.g.,
    \begin{itemize}
        \item "I hate Hilary Clinton"
        \item "She is so lame"
        \item "Donald Trump is a bellend"
    \end{itemize}

    \item Entries abusive of other protected characteristics, but not gender, e.g.,
    \begin{itemize}
        \item "Jews make me sick"
        \item "White honkies gona dieeeeee if they cross me"
    \end{itemize}

    NB: Abusive entries that attack gender with other characteristics (e.g., “I hate black women”), or contain gendered slurs (e.g., “Don’t be such a bitch”) should still be labelled Sexist

    \item Entries that criticise feminism as a theoretical framework, ideology or practice, e.g.,
    \begin{itemize}
        \item "I dont identify as a feminist. I just dont like the connotations, I try not to be political."
        \item "Feminism isnt a well formulated theory, it’s not disprovable and so isnt a proper science."
    \end{itemize}
    NB: Take care to distinguish between criticism of feminism as a theory, which by itself should be labelled Not Sexist, and abuse of feminists as people (e.g., "Feminists are such loony eyed old bags"), which should be labelled Sexist. However, entries which combine criticism of feminism with abuse of feminists should be labelled Sexist.
\end{enumerate}

\section{Input Format and Examples}
\label{app:input}

We provide examples for five prompt templates in Table \ref{tab:prompt}. The example text ``So tell single moms to stop raising rapists.'' and its label is ``A. GBV''.

\section{Experimental Details}
\label{app:experiment}

\paragraph{Models} We implement three models in \autoref{sec:experiments} based on the Python library Transformers provided by Hugging Face \citep{wolf-etal-2020-transformers}. 
These models are pre-trained and available in Hugging Face models, namely \texttt{roberta-base},  
\texttt{cardiffnlp/twitter-roberta-base-h- ate-latest}, \texttt{google/flan-t5-xl}, \texttt{meta-llama/Llama-2-7b-hf}, \texttt{meta-llama/Meta-Llama-3-8B}, and \texttt{mistralai/Mistral-7B-v0.1}.

\paragraph{Experimental Setting} 
We randomly split our training set into training (4,800 samples) and validation (1,200 samples) sets by the ratio of 4:1 for fine-tuning.
For ICL experiments, we use OpenPrompt \citep{ding2022openprompt}, a standard framework for prompt learning over pre-trained language models. We use the default hyperparameters in Hugging Face. 
For fine-tuning LLMs, we prioritise several hyperparameters, using cross-entropy loss and the AdamW optimiser \citep{loshchilov2018decoupled} with a $1e-3$ weight decay. We use the learning rate $1e-3$ for \texttt{FLAN-T5} and $3e-4$ for the others. We set the batch size to 128, the micro batch size to 4, the maximum sequence length to 512.
We use 5 epochs for \texttt{LLaMA 2} and \texttt{LLaMA 3}, 4 epochs for \texttt{Mistral} and 2 epochs for \texttt{FLAN-T5}.
For the LoRA hyperparameters, we set the attention dimension to 8, the alpha parameter to 16, and the dropout value to 0.05 for LoRA layers. We use the task type ``SEQ\_2\_SEQ\_LM'' and modules q and v for \texttt{FLAN-T5}, and use ``CAUSAL\_LM'' and modules q\_proj and v\_proj for the other decoder-only LLMs.

We save the checkpoint with the highest F1 score as the final model.


\paragraph{Computation} All experiments are conducted on the Heriot-Watt University high-performance computing facility (DMOG) and the Edinburgh International Data Facility (EIDF).

\begin{table*}[ht!]
    \centering
    \small
    \begin{tabular}{l}
    \toprule 
     \textbf{Label prompt:} \\\texttt{<text>} This text is classified as \texttt{<label>}\\
     \textbf{Example:} \\So tell single moms to stop raising rapists. This text is classified as \\ \toprule 
    
    \textbf{Task description:} \\\texttt{<task description>} Text: \texttt{<text>} \texttt{<choices>} Answer: \\
     \textbf{Example:}  \\Classify the following message from a social media platform. It might contain a form of gender-based violence \\(GBV). Output A. GBV if it contains GBV, or B. Not GBV if not. \\Text: So tell single moms to stop raising rapists. \\Choices: A. GBV or B. Not GBV. \\Answer: \\ \toprule 
     
     \textbf{Few-shot demonstrations:} \\\texttt{<task description> <demonstration>} Text: \texttt{<text>} \texttt{<choices>} Answer:\\ 
     \textbf{Example:}  \\Classify the following message from a social media platform. It might contain a form of gender-based violence \\(GBV). Output A. GBV if it contains GBV, or B. Not GBV if not. 
     \\Text: Fuck the niggers and the jews. Both have a history of eating white people and fuck you for suggesting I \\should get along with people who want to eat me. \\ Choices: A. GBV or B. Not GBV.\\Answer: B. Not GBV
     \\Text: Well then good because someone has to knock her ass back in line when she acts like a woman and posts \\bullshit about her (((based minorities))). If someone is not white they are NOT one of us. \\ Choices: A. GBV or B. Not GBV. \\Answer: A. GBV 
     \\Text: So tell single moms to stop raising rapists. \\Choices: A. GBV or B. Not GBV.\\Answer: \\ \toprule 
     
     \textbf{Annotator descriptions (short):} \\\texttt{<task description> <annotator description>} Text: \texttt{<text>} \texttt{<choices>} Answer: \\ 
     \textbf{Example:} \\Classify the following message from a social media platform. It might contain a form of gender-based violence \\(GBV). Output A. GBV if it contains GBV, or B. Not GBV if not.
     \\This annotator is a 32-year-old asian female, who is heterosexual and right-wing/conservative politics.\\Three scales are used to show the annotator's attitudes, namely the Very Short Authoritarianism (VSA) scale to \\measure Right Wing Authoritarianism (RWA), the Short Social Dominance Orientation (SSDO) scale to measure \\Social Dominance Orientation (SDO), and the Brief Hostile Neosexism scale to measure Hostile Neosexism. \\This worker is moderate right wing authoritarianism, moderate social dominance orientation, and moderate hostile \\neosexism. 
     \\Text: So tell single moms to stop raising rapists.\\Choices: A. GBV or B. Not GBV.\\ Answer:  \\ \toprule 
     
     \textbf{Annotator descriptions (full):} \\\texttt{<task description> <annotator description>} Text: \texttt{<text>} \texttt{<choices>} Answer: \\ 
     \textbf{Example:} \\Classify the following message from a social media platform. It might contain a form of gender-based violence \\(GBV). Output A. GBV if it contains GBV, or B. Not GBV if not. 
     \\This annotator is a 32-year-old asian female, who is heterosexual and right-wing/conservative politics.\\Three scales are used to show the annotator's attitudes, namely the Very Short Authoritarianism (VSA) scale to \\measure Right Wing Authoritarianism (RWA), the Short Social Dominance Orientation (SSDO) scale to measure \\Social Dominance Orientation (SDO), and the Brief Hostile Neosexism scale to measure Hostile Neosexism. \\This worker is moderate right wing authoritarianism, moderate social dominance orientation, and moderate hostile \\neosexism.  The following are questions\\ and annotator's answers \\from each scale.\\Scale 1: Very Short Authoritarianism Scale (VSA)\\Statement 1: It\u2019s great that many young people today are prepared to defy authority.\\Answer: Strongly agree\\Statement 2: What our country needs most is discipline, with everyone following our leaders in unity.\\Answer: Strongly agree\\Statement 3: God\u2019s laws about abortion, pornography, and marriage must be strictly followed before it is too late.\\Answer: Very strongly agree\\Statement 4: There is nothing wrong with premarital sexual intercourse.\\Answer: Unsure or neutral\\Statement 5: Our society does NOT need tougher Government and stricter Laws.\\Answer: Strongly agree\\Statement 6: The facts on crime and the recent public disorders show we have to crack down harder on troublemakers, \\if we are going to preserve law and order.\\Answer: Strongly agree\\Scale 2: Short Social Dominance Orientation Scale (SSDO)\\Statement 1: In setting priorities, we must consider all societal groups.\\Answer: Strongly disagree
     \\Statement 2: We should not push for equality of societal groups.\\Answer: Slightly disagree
     \\\toprule 
    \end{tabular}
    \label{tab:prompt}
\end{table*}

\begin{table*}[ht!]
    \centering
    \small
    \begin{tabular}{l}
    [Continued Table]\\
    \toprule 

     Statement 3: The equality of societal groups should be our goal.
     \\Answer: Strongly agree
     \\Statement 4: Superior societal groups should dominate inferior groups.\\Answer: Strongly disagree\\Scale 3: Brief Hostile Neosexism Scale\\Statement 1: Some of the demands of the feminist movement seem to me to be a bit exaggerated.\\Answer: Somewhat agree\\Statement 2: I sometimes feel that our society pays too much attention to the rights of certain minorities.\\Answer: Strongly agree\\Statement 3: In the name of equality, many women try to gain certain privileges.\\Answer: Somewhat agree\\Statement 4: Many women interpret innocent comments and actions as sexist.\\Answer: Slightly disagree\\Statement 5: Women are easily offended.\\Answer: Somewhat disagree\\Statement 6: Women exaggerate the problems they suffer because they are women.\\Answer: Slightly disagree
     \\Text: So tell single moms to stop raising rapists.\\Choices: A. GBV or B. Not GBV.\\ Answer:  \\ 
     \toprule 
     
     \textbf{Combined prompt (short annotator description):} \\\texttt{<task description> <demonstration>} \texttt{<annotator description>} Text: \texttt{<text>} \texttt{<choices>} \\Answer: \\
     \textbf{Example:}  \\Classify the following message from a social media platform. It might contain a form of gender-based violence \\(GBV). Output A. GBV if it contains GBV, or B. Not GBV if not.  
     \\This annotator is a 32-year-old white female, who is bisexual and left-wing/liberal politics.\\Three scales are used to show the annotator's attitudes, namely the Very Short Authoritarianism (VSA) scale to\\ measure Right Wing Authoritarianism (RWA), the Short Social Dominance Orientation (SSDO) scale to measure\\ Social Dominance Orientation (SDO), and the Brief Hostile Neosexism scale to measure Hostile Neosexism. This \\worker is moderate right wing authoritarianism, high social dominance orientation, and high hostile neosexism.
     \\Text: Fuck the niggers and the jews. Both have a history of eating white people and fuck you for suggesting I should \\get along with people who want to eat me. \\ Choices: A. GBV or B. Not GBV.\\Answer: A. GBV 
     \\This annotator is a 48-year-old asian female, who is heterosexual and right-wing/conservative politics.\\Three scales are used to show the annotator's attitudes, namely the Very Short Authoritarianism (VSA) scale to\\ measure Right Wing Authoritarianism (RWA), the Short Social Dominance Orientation (SSDO) scale to measure \\Social Dominance Orientation (SDO), and the Brief Hostile Neosexism scale to measure Hostile Neosexism. This \\worker is moderate right wing authoritarianism, moderate social dominance orientation, and high hostile neosexism.
     \\Text: Fuck the niggers and the jews. Both have a history of eating white people and fuck you for suggesting I should\\ get along with people who want to eat me. \\ Choices: A. GBV or B. Not GBV.\\Answer: B. Not GBV

     \\ This annotator is a 35-year-old american male, who is bisexual and right-wing/conservative politics.\\Three scales are used to show the annotator's attitudes, namely the Very Short Authoritarianism (VSA) scale to \\measure Right Wing Authoritarianism (RWA), the Short Social Dominance Orientation (SSDO) scale to measure\\ Social Dominance Orientation (SDO), and the Brief Hostile Neosexism scale to measure Hostile Neosexism. This\\ worker is moderate right wing authoritarianism, moderate social dominance orientation, and high hostile neosexism.
     \\Text: Well then good because someone has to knock her ass back in line when she acts like a woman and posts \\bullshit about her (((based minorities))). If someone is not white they are NOT one of us. \\ Choices: A. GBV or B. Not GBV. \\Answer: A. GBV 
     \\ This annotator is a 32-year-old asian female, who is heterosexual and right-wing/conservative politics.\\Three scales are used to show the annotator's attitudes, namely the Very Short Authoritarianism (VSA) scale to \\measure Right Wing Authoritarianism (RWA), the Short Social Dominance Orientation (SSDO) scale to measure \\Social Dominance Orientation (SDO), and the Brief Hostile Neosexism scale to measure Hostile Neosexism. This\\ worker is moderate right wing authoritarianism, moderate social dominance orientation, and moderate hostile\\ neosexism.
     \\Text: Well then good because someone has to knock her ass back in line when she acts like a woman and posts \\bullshit about her (((based minorities))). If someone is not white they are NOT one of us. \\ Choices: A. GBV or B. Not GBV. \\Answer: B. Not GBV
     
     \\ Text: So tell single moms to stop raising rapists.\\Choices: A. GBV or B. Not GBV.\\ Answer: \\ \toprule 
    \end{tabular}
    \caption{Input examples with different prompt templates.}
    \label{tab:prompt}
\end{table*}

\section{Experimental Results}
\label{app:experiment-result}

We provide comprehensive results for ICL and fine-tuning experiments in the following tables.
\autoref{tab:experiment-icl-single-flant5-full} shows more ICL results on two subsets based on \autoref{tab:experiment-icl-single-flant5}. \autoref{tab:experiment-ft-single-flant5}, \autoref{tab:experiment-ft-single-llama2}, \autoref{tab:experiment-ft-single-llama3}, and \autoref{tab:experiment-ft-single-mistral} present more fine-tuning results among LLMs tested on original labels by extending \autoref{tab:experiment-ft-single}.

\begin{table*}[ht!]
    \centering
    \small
    \begin{tabular}{l|c|c|c|c|c|c|c|c|c|}
        \textbf{Model: \texttt{FLAN-T5}} & \multicolumn{3}{c|}{\textbf{Original Label - GBV}} & \multicolumn{3}{c|}{\textbf{Original Label - Sexist}} & \multicolumn{3}{c|}{\textbf{Original Label - Misogynistic}}\\
         & \texttt{Our} & \texttt{EDOS} & \texttt{DOM} & \texttt{Our} & \texttt{EDOS} & \texttt{DOM} & \texttt{Our} & \texttt{EDOS} & \texttt{DOM} \\
         \toprule 
        Majority class (single) & $44.54$ & $43.61$ & $47.15$ & $44.54$ & $43.61$ & $47.15$ & $44.54$ & $43.61$ & $47.15$\\
        task & $63.60$ & $62.43$ & $67.09$ & $66.94$ & $66.32$ &$64.14$ & $71.47$ & $71.52$ & $67.21$\\
        +demos & $65.49$ & $64.57$ & $67.21$ & $69.19$ & $68.82$ & $67.21$ & $\mathbf{72.84}$ & $\mathbf{73.16}$ & $67.69$\\
        +anno (short) & $65.72$ & $64.84$ & $67.09$ & $63.40$ & $62.28$ & $62.89$ & $67.63$ & $66.15$ & $\mathbf{69.48}$ \\
        +anno (short)+demos & $65.86$ & $65.69$ & $63.62$ & $\mathbf{70.21}$ & $\mathbf{69.46}$ & $\mathbf{70.00}$ & $70.45$ & $70.32$& $67.81$\\
        +anno (full) & $\mathbf{67.20}$ & $\mathbf{66.51}$ & $\mathbf{67.78}$ & $67.32$ & $66.48$ & $65.75$ & $69.51$ & $69.44$ & $67.09$ \\
        +anno (full)+demos & $\mathbf{67.20}$ & $\mathbf{66.51}$ & $\mathbf{67.78}$ & $67.32$ & $66.48$ & $65.75$ & $69.51$ & $69.44$ & $67.09$\\ \toprule 
        &  \multicolumn{3}{c|}{\textbf{New Label - GBV}} & \multicolumn{3}{c|}{\textbf{New Label - Sexist}} & \multicolumn{3}{c|}{\textbf{New Label - Misogynistic}}   \\ \toprule 
        Majority class (single) & $36.14$ & $35.71$ & $37.41$  & $36.14$ & $35.71$ & $37.41$  & $36.14$ & $35.71$ & $37.41$\\
        task  & $60.29$ & $60.53$ & $56.81$ & $65.25$ & $\mathbf{64.87}$ & $62.04$ & $61.12$ &$61.28$ & $56.95$\\
        +demos & $\mathbf{62.60}$ & $\mathbf{62.92}$ & $\mathbf{58.49}$ & $63.11$ & $64.24$ & $54.39$ & $59.67$ & $60.18$ & $54.25$ \\
        +anno (short) & $61.13$ & $61.35$ & $57.05$ & $\mathbf{65.62}$ & $64.28$ & $\mathbf{66.41}$ & $\mathbf{64.69}$ & $\mathbf{64.00}$ & $\mathbf{62.15}$\\
        +anno (short)+demos & $59.91$ & $60.50$ & $54.80$ & $63.43$ & $63.72$ & $53.88$ & $62.32$ & $62.29$ & $59.32$\\
        +anno (full)  & $61.22$ &  $61.38$ & $57.65$ & $62.23$ & $62.21$ & $56.67$ & $58.20$ & $58.57$ & $53.97$\\
        +anno (full)+demos  & $61.22$ & $61.38$ & $57.65$ & $62.23$ & $62.21$ & $56.67$ &  $58.20$ & $58.57$ & $53.97$\\ 
    \end{tabular}
    \caption{Results of in-context learning on \texttt{FLAN-T5} by using different label texts: (i) ``GBV'' as the aggregated label, (ii) ``Sexist'' from \texttt{EDOS} dataset, and (iii) ``Misogynistic'' from \texttt{DOM} dataset. Six different input prompts are evaluated among three label texts. The best results are shown in bold by column.}
    \label{tab:experiment-icl-single-flant5-full}
\end{table*}

\begin{table*}[ht!]
    \centering
    \small
    \begin{tabular}{l|c|c|c|}
         \textbf{Model: \texttt{FLAN-T5}} & \multicolumn{3}{c|}{\textbf{Original Label}} \\
         & \texttt{All} & \texttt{EDOS} & \texttt{DOM} \\
         \toprule 
        Majority class (single) & $44.54\pm0.0$ & $43.61\pm0.0$ & $47.15\pm0.0$ \\
        task &  $66.82\pm1.03$ & $64.94\pm1.23$ & $71.36\pm2.54$\\
        +demos & $67.06\pm1.35$ & $66.61\pm2.16$ & $67.68\pm1.58$ \\
        +anno (short) &  $68.19\pm2.62$ & $66.78\pm2.41$ & $71.31\pm2.79$ \\
        +anno (short)+demos &  $67.94\pm1.77$ & $64.98\pm1.68$ & $70.83\pm0.65$ \\
        +anno (full) &  $71.51\pm1.47$ & $70.24\pm0.85$ & $73.32\pm1.74$ \\
        +anno (full)+demos &  $70.04\pm0.33$ & $69.47\pm0.69$ & $72.64\pm0.23$ \\ \toprule 
        & \multicolumn{3}{c|}{\textbf{New Label}} \\ \toprule 
         Majority class (single) & $36.14\pm0.0$ & $35.71\pm0.0$ & $37.41\pm0.0$ \\
        task &  $63.78\pm1.84$ & $64.06\pm1.73$ & $62.96\pm3.06$\\
        +demos & $65.12\pm1.66$ & $65.33\pm0.94$ & $64.69\pm2.44$ \\
        +anno (short) &  $65.79\pm1.89$ & $66.53\pm0.77$ & $64.37\pm2.11$ \\
        +anno (short)+demos &  $64.95\pm1.03$ & $65.06\pm0.45$ & $64.02\pm1.69$ \\
        +anno (full) & $64.50\pm1.17$ & $64.88\pm1.73$ & $63.11\pm2.07$\\
        +anno (full)+demos &  $62.23\pm0.54$ & $63.07\pm0.60$ & $59.90\pm1.76$ \\
    \end{tabular}
    \caption{Results of fine-tuning \texttt{FLAN-T5} on single annotator labels using different input prompts for the GBV detection task. }
    \label{tab:experiment-ft-single-flant5}
\end{table*}

\begin{table*}[ht!]
    \centering
    \small
    \begin{tabular}{l|c|c|c|}
         \textbf{Model: \texttt{LLaMA 2}} & \multicolumn{3}{c|}{\textbf{Original Label}} \\
         & \texttt{All} & \texttt{EDOS} & \texttt{DOM} \\
         \toprule 
        Majority class (single) & $44.54\pm0.0$ & $43.61\pm0.0$ & $47.15\pm0.0$ \\
        task &  $49.32\pm2.57$ & $49.53\pm1.29$ & $47.30\pm7.95$\\
        +demos & $47.09\pm1.45$ & $48.59\pm1.92$ & $42.46\pm2.04$\\
        +anno (short) &  $52.06\pm1.17$ & $51.89\pm2.00$ & $51.63\pm2.19$ \\
        +anno (short)+demos &  $51.12\pm1.98$ & $51.29\pm3.08$ & $49.98\pm1.78$ \\
        +anno (full) &  $52.23\pm1.76$ & $52.05\pm1.95$ & $51.66\pm1.08$ \\
        +anno (full)+demos &  $52.09\pm1.38$ & $50.90\pm2.12$ & $55.96\pm1.83$ \\ \toprule 
        & \multicolumn{3}{c|}{\textbf{New Label}} \\ \toprule 
         Majority class (single) &$36.14\pm0.0$ & $35.71\pm0.0$ & $37.41\pm0.0$  \\
        task &  $51.87\pm1.76$ & $51.45\pm2.04$ & $52.43\pm2.39$\\
        +demos & $49.40\pm1.79$ & $50.36\pm1.56$ & $46.53\pm2.39$\\
        +anno (short) &  $51.17\pm1.58$ & $50.20\pm0.83$ & $53.95\pm4.11$ \\
        +anno (short)+demos &  $41.16\pm1.06$ & $41.59\pm1.29$ & $37.13\pm1.42$ \\
        +anno (full) &  $53.21\pm0.12$ & $51.60\pm1.07$ & $57.76\pm3.86$ \\
        +anno (full)+demos & $50.02\pm1.10$ & $47.18\pm1.36$ & $59.55\pm0.24$ \\
    \end{tabular}
    \caption{Results of fine-tuning \texttt{LLaMA 2} on single annotator labels using different input prompts for the GBV detection task. }
    \label{tab:experiment-ft-single-llama2}
\end{table*}

\begin{table*}[ht!]
    \centering
    \small
    \begin{tabular}{l|c|c|c|}
         \textbf{Model: \texttt{LLaMA 3}} & \multicolumn{3}{c|}{\textbf{Original Label}} \\
         & \texttt{All} & \texttt{EDOS} & \texttt{DOM} \\
         \toprule 
        Majority class (single) & $44.54\pm0.0$ & $43.61\pm0.0$ & $47.15\pm0.0$ \\
        task &  $47.92\pm0.84$ & $48.55\pm0.75$ & $45.01\pm1.72$\\
        +demos & $50.04\pm2.12$ & $49.42\pm1.74$ & $51.27\pm1.88$\\
        +anno (short) &  $47.23\pm2.32$ & $45.45\pm1.66$ & $53.91\pm6.25$ \\
        +anno (short)+demos &  $41.16\pm1.06$ & $41.59\pm1.29$ & $37.13\pm1.42$ \\
        +anno (full) &  $50.65\pm0.27$ & $50.33\pm0.30$ & $51.09\pm0.68$ \\
        +anno (full)+demos &  $45.07\pm0.49$ & $44.40\pm0.68$ & $47.27\pm0.19$ \\ \toprule 
        & \multicolumn{3}{c|}{\textbf{New Label}} \\ \toprule 
         Majority class (single) &$36.14\pm0.0$ & $35.71\pm0.0$ & $37.41\pm0.0$  \\
        task &  $50.32\pm2.75$ & $50.68\pm2.53$ & $48.65\pm4.59$\\
        +demos & $52.12\pm1.05$ & $52.07\pm1.37$ & $51.70\pm1.26$\\
        +anno (short) &  $43.39\pm1.47$ & $41.57\pm0.59$ & $49.29\pm6.43$ \\
        +anno (short)+demos &  $50.40\pm0.53$ & $50.40\pm0.27$ & $48.59\pm1.98$ \\
        +anno (full) & $51.70\pm2.51$ & $51.17\pm2.36$ & $53.12\pm3.03$\\
        +anno (full)+demos &  $43.31\pm0.29$ & $40.70\pm0.73$ & $51.57\pm1.16$ \\
    \end{tabular}
    \caption{Results of fine-tuning \texttt{LLaMA 3} on single annotator labels using different input prompts for the GBV detection task. }
    \label{tab:experiment-ft-single-llama3}
\end{table*}

\begin{table*}[ht!]
    \centering
    \small
    \begin{tabular}{l|c|c|c|}
         \textbf{Model: \texttt{Mistral}} & \multicolumn{3}{c|}{\textbf{Original Label}} \\
         & \texttt{All} & \texttt{EDOS} & \texttt{DOM} \\
         \toprule 
        Majority class (single) & $44.54\pm0.0$ & $43.61\pm0.0$ & $47.15\pm0.0$ \\
        task &  $58.75\pm0.78$ & $59.42\pm0.96$ & $54.86\pm1.14$\\
        +demos & $44.51\pm0.94$ & $43.61\pm0.27$ & $47.03\pm0.59$ \\
        +anno (short) & $45.90\pm1.33$ & $47.07\pm1.64$ & $41.20\pm1.81$ \\
        +anno (short)+demos & $64.31\pm0.45$ & $66.85\pm0.31$ & $54.51\pm0.77$\\
        +anno (full) & $44.41\pm1.19$ & $43.52\pm2.05$ & $46.92\pm1.72$\\
        +anno (full)+demos & $47.74\pm0.66$ & $48.15\pm0.41$ & $43.66\pm0.63$\\ \toprule 
        & \multicolumn{3}{c|}{\textbf{New Label}} \\ \toprule 
         Majority class (single) & $36.14\pm0.0$ & $35.71\pm0.0$ & $37.41\pm0.0$ \\
        task &  $59.20\pm2.10$ & $57.45\pm1.18$ & $64.79\pm1.54$\\
        +demos & $41.07\pm1.18$ & $39.86\pm1.11$ & $44.92\pm2.31$\\
        +anno (short) & $52.56\pm1.79$& $53.98\pm0.69$ & $57.61\pm1.36$ \\
        +anno (short)+demos & $67.40\pm1.55$& $67.53\pm0.93$ & $66.62\pm1.10$ \\
        +anno (full) & $40.96\pm2.97$ & $40.23\pm2.12$ & $42.92\pm1.87$ \\
        +anno (full)+demos & $54.15\pm0.46$ & $54.00\pm0.77$ & $52.83\pm2.05$\\
    \end{tabular}
    \caption{Results of fine-tuning \texttt{Mistral} on single annotator labels using different input prompts for the GBV detection task. }
    \label{tab:experiment-ft-single-mistral}
\end{table*}
\end{document}